\documentclass[journal]{IEEEtran}
\usepackage{times}
\usepackage{epsfig}
\usepackage{graphicx}
\usepackage{amsmath}
\usepackage{amssymb}
\usepackage{subfigure}
\usepackage{cite}

\ifCLASSINFOpdf
\else
\fi

\begin{document}
%
\title{Automatic Segmentation of Dynamic Objects from an Image Pair}
%
%
%

\author{Sri Raghu Malireddi
        and Shanmuganathan Raman
\thanks{Sri Raghu Malireddi and Shanmuganathan Raman are with the Electrical Engineering \& Computer Science and Engineering, Indian Institute of Technology Gandhinagar, Palaj,
GJ, 382355 India. e-mail: (sriraghu\_malireddi, shanmuga)@iitgn.ac.in}\\Electrical Engineering \& Computer Science and Engineering, Indian Institute of Technology Gandhinagar}
\maketitle

\begin{abstract}
Automatic segmentation of objects from a single image is a challenging problem which generally requires training on large number of images. We consider the problem of automatically segmenting only the dynamic objects from a given pair of images of a scene captured from different positions. We exploit dense correspondences along with saliency measures in order to first localize the interest points on the dynamic objects from the two images. We propose a novel approach based on techniques from computational geometry in order to automatically segment the dynamic objects from both the images using a top-down segmentation strategy. We discuss how the proposed approach is unique in novelty compared to other state-of-the-art segmentation algorithms. We show that the proposed approach for segmentation is efficient in handling large motions and is able to achieve very good segmentation of the objects for different scenes. We analyse the results with respect to the manually marked ground truth segmentation masks created using our own dataset and provide key observations in order to improve the work in future.
\end{abstract}

\begin{IEEEkeywords}
Segmentation, Dense Correspondence, Saliency
\end{IEEEkeywords}

%
\IEEEpeerreviewmaketitle

\section{Introduction}
Real world scenes are dynamic where most objects constantly change their positions over time. Images of the scenes contain only the 2D projections of the actual motion field of the objects. Segmenting the objects in motion from images provide us much better understanding of the 3D scene. This information is generally obtained by assuming that the camera is static with respect to the scene and the object motion is not very large. In this scenario, optical flow field provides us information regarding the 2D motion field corresponding to the moving objects \cite{horn1981determining}. There are efficient background subtraction methods which enable us to segment the objects which are dynamic \cite{stauffer1999adaptive}.

Consider the setting in which we have two images of a dynamic scene captured using the same camera in two different positions. This scenario is quite common due to the proliferation of digital cameras in mobile devices. These images can be considered as stereo images of the scene with each camera capturing the scene at a different time instant. Given two such images of the scene, we would like to segment only the moving objects from the images. This problem is challenging as the object could have moved to any position in the scene or even could have moved out of the field of view. The perspective projection on to the image plane would also be different for the two images. With no additional information, we would like to perform automatic segmentation of the objects from the given two images.

The primary motivation behind this work is to use the motion information present only in the dynamic regions of the scene in order to perform segmentation. The first step to achieve this task is to separate the static and dynamic regions from the given two images. We achieve this by estimating dense correspondence between the two images and using the vector connecting the corresponding coordinates. The dynamic objects are localized by the corresponding saliency measures along with the motion information. The next step is to find the interest points on the dynamic objects present in the scene and to separate the interest points of different dynamic objects. This is achieved through clustering operation.

We then need to use the interest points present on the dynamic objects as the seed in order to perform complete segmentation of the object. In practice, such points are sparsely distributed on the object and therefore not suitable for use directly. We generate a convex hull which covers all the interest points present in an object and estimate a minimum area bounding rectangle covering the convex hull. This rectangle is then used to segment the object using a top-down approach for segmentation. This enables us to segment out multiple dynamic objects present in the scene in both the images. 

The primary contributions of this paper are listed below.

\begin{enumerate}
\item Given two images of a dynamic scene captured at various time instants, we have developed an approach to segment automatically all the dynamic objects in the scene.
\item We show that segmentation is possible even when camera positions are vastly different from each other while capturing the two images.
\item We show that the segmentation is very effective even when dynamic objects are allowed to undergo large and non-rigid motions.
\end{enumerate}

We have excellent single image segmentation algorithms presently available in literature which require human interaction (\cite{rother2004grabcut}, \cite{vicente2008graph}). The main idea behind the work is that we are given extra information in the form of another image of the same scene shot by the same or different camera from a different position and time. We would like to develop an automatic image segmentation method using this extra information available in order to segment out the dynamic objects present in the scene. The proposed novel approach can aid in computer vision problems such as object recognition and visual tracking. Our work is the first of its kind to completely segment multiple dynamic objects from two images captured using a hand-held camera even when there is a significant amount of background clutter to the best of our knowledge.

The rest of the paper is organized as follows. Section 2 describes some of the works in the recent past which are related to the techniques used in the proposed approach. Section 3 presents the various steps involved in the process of segmenting dynamic objects present in the scene from the two images. We discuss the importance of each processing step involved in the present work. Section 4 provides a discussion on the results obtained along with the discussion of the segmentation results for different challenging scenes. We show that the proposed framework is able to generate excellent segmentation of the moving objects in most of the scenes. Section 5 provides the conclusion along with some suggestions for the future enhancement of the present work based on the observations from the results obtained.

\section{Related Work}
The problem of estimating the motion from a given set of images of a dynamic scene have been studied for static camera case using optical flow techniques \cite{horn1986robot}. When the camera position is different for the two images captured, one treats the problem as a stereo imaging problem assuming static scene (\cite{marr1976cooperative}, \cite{scharstein2002taxonomy}, \cite{longuet1981computer}). These classical problems have been of interest to the computer vision communication over the past four decades. Optical flow techniques fail when there is large motion in the scene while the stereo algorithms work only for static scenes. We shall start this section with a survey of major segmentation approaches followed by segmentation methods.
\subsection{Segmentation}
Image segmentation has always been one of the most challenging problems in computer vision. Some of the earlier approaches for image segmentation used active contours \cite{kass1988snakes}. These approaches required some initialization of the curve to evolve later. A closely related approach to image segmentation is the use of level set. An overview of level set segmentation approaches can be found in \cite{cremers2007review}. Another class of segmentation algorithms use Mumford-Shah functional in the variational framework \cite{mumford1989optimal}. One of the examples of this approach uses curve evolution to achieve image segmentation \cite{tsai2001curve}. The approaches described above are top down in the sense that they start with a rough segmentation and try to refine it over multiple iterations in order to obtain proper segmentation.

Another class of segmentation algorithms consider image in the discrete domain as a graph and try to optimally segment desired regions from the image. One of the earliest methods based on graphs is the normalized cuts by Shi and Malik \cite{shi2000normalized}. This algorithm laid the foundation for over-segmentation approaches which opened up research on bottom-up segmentation through grouping \cite{ren2003learning}. Another instance of the use of image as a graph in order to segment can be found in \cite{felzenszwalb2004efficient}. These approaches work on the common objective that the nodes of the graph which are similar must be grouped together.

There are methods using graphs based on max-flow min-cut algorithms which are popularly known as graph cuts \cite{boykov2001fast}. The energy function in a graph can be minimized by using appropriate partitioning of the graphs through cuts on the edges which link vertices which are dissimilar \cite{kolmogorov2004energy}. An interactive approach based on the initialisation of a bounding box on the image was proposed using graph cuts in order to obtain segmentation. This method is popularly known as `GrabCut'  \cite{rother2004grabcut}. A more advanced version of the algorithm which makes use of Djikstra algorithm for segmentation of thin structures was proposed in \cite{vicente2008graph}. An overview of many other undirected graph (Markov random field) based image segmentation can be found in the book by Blake \emph{et al.} \cite{blake2011markov}.

Due to the recent proliferation of internet images, researchers have turned their attention to co-segment common objects present in multiple images. This objective can be addressed in a supervised learning framework  \cite{batra2010icoseg} or in a completely unsupervised framework \cite{rubinstein2013unsupervised}. A recent approach for motion segmentation from video sequence can be found in \cite{ochs2014segmentation}. This work assumes that the frames are captured within a fraction of a second and does not apply to scenes which are captured with a delay of the order of seconds.  The recent work tries to segment objects from two images of the scene captured with seconds of gap. This work is the closest recent attempt to the problem we address \cite{poling2014new}. This work proposes a new technique called global dimension reduction in order to achieve this objective. However, this method only estimates some points on the dynamic object and does not provide complete segmentation of the object.

\subsection{Dense Correspondence}
We shall now review some of the works related to estimation of dense correspondence which is used to match two images of a scene which has undergone significant changes. The estimation of accurate nearest neighbour for matching is computationally intractable. Approximate algorithms for estimating the nearest neighbour and thereby match two images was first proposed in  \cite{arya1994optimal}. Considering image as a collection of patches, a faster algorithm to compute one nearest neighbour patch to a given patch called PatchMatch was proposed in \cite{barnes2009patchmatch}. This approach was extended and generalized to obtain multiple nearest neighbours for a given patch in the following work  \cite{barnes2010generalized}. An approach called coherency sensitive hashing improved the idea of locality sensitive hashing in order to compute the approximate nearest neighbours much faster than PatchMatch \cite{korman2011coherency}.

The above methods for dense correspondence were not able to handle non-rigid motion of the objects in the scene effectively. An algorithm based on matching of scale invariant feature transform (SIFT) features between two images called SIFT-flow has been proposed to take into account such complex motions \cite{liu2011sift}. Another method called non-rigid dense correspondence is specifically meant to match features between two images of a scene where the objects have undergone non-rigid motion. This approach called non-rigid dense correspondence (NRDC) is shown to be more accurate compared to SIFT-flow  \cite{hacohen2011non}. Another approach for the dense correspondence of deformable objects is proposed in  \cite{kim2013deformable}. Dense correspondences have a number of applications such as image retargeting \cite{simakov2008summarizing}, high dynamic range image reconstruction (\cite{sen2012robust}, \cite{hu2013hdr}),  and image melding \cite{darabi2012image}.   A recent work which has been accepted to be published builds on these ideas and develops a framework to accurately match features between two images having common object present in different orientations \cite{dekel2015}.

\subsection{Saliency Estimation}
Visual saliency refers to the attention a human being has on some particular object in the scene. Visual attention was first proposed and models were built in the seminal work by Itti \emph{et al.}
\cite{itti1998model}. The interest towards estimating the saliency from an image grew in the computer vision community. A recent approach exploits global contrast in order to produce a saliency map of a given image \cite{cheng2011global}. Another recent work involves measuring saliency from the context in a given image \cite{goferman2012context}. These approaches for saliency detection work at a pixel. A more efficient and accurate saliency detection approach involves estimating saliency at the patch level \cite{margolin2013makes}. By processing the patch corresponding to a given pixel location one can produce a good saliency map with this approach.
\section{Proposed Approach}
In this section, we provide the detailed explanation of the various steps involved in the proposed approach for segmentation. As already discussed, the challenge is to segment the dynamic objects in the scene from two images even when the camera position has changed while capturing them. Two such images are shown in Figure \ref{Fig:1} for illustration. We shall provide the intuitive reasoning behind the choice of different techniques used in this work in each subsection.


\subsection{Saliency Guided Dense Correspondence}
\begin{figure}[ht]
\centering
\subfigure[]{\includegraphics[width=0.23\textwidth]{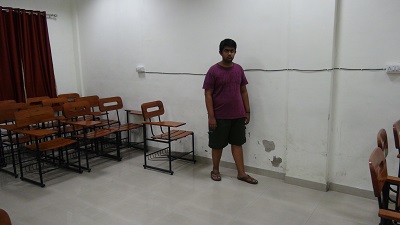}}
\subfigure[]{\includegraphics[width=0.23\textwidth]{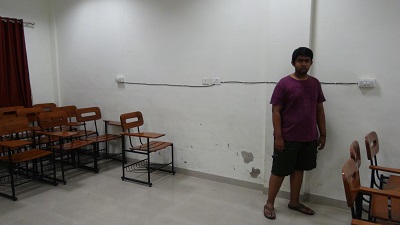}}
\caption{Two images of a dynamic scene captured using a hand-held camera.} 
\label{Fig:1}
\end{figure}

\begin{figure}[ht]
\centering
\subfigure[]{\includegraphics[width=0.23\textwidth]{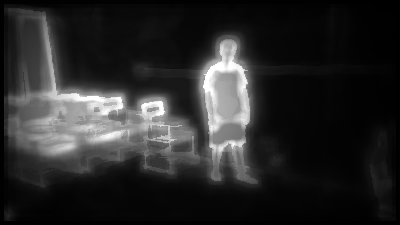}}
\subfigure[]{\includegraphics[width=0.23\textwidth]{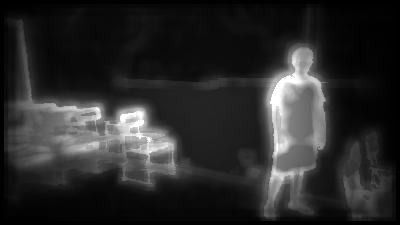}}
\caption{Saliency maps for the images shown in Figure \ref{Fig:1}.} 
\label{Fig:2}
\end{figure}

Given two images of a dynamic scene captured using a camera from different positions, we would like to first estimate the salient regions in both the images. This is due to the fact that the salient regions contain the objects of interest which we would like to segment. The saliency map in both the images will contain all the objects present in the scene, not exclusively the objects which are dynamic. We segment the regions in both the images using saliency measure and a threshold in order to better localize the objects of our interest. This is achieved by creating a binary mask out of the saliency map and multiplying it with the image. We use the approach by Margolin \emph{et al.} for computing the saliency map \cite{margolin2013makes}. The saliency maps estimated for the images shown in Figure \ref{Fig:1} are depicted in Figure \ref{Fig:2}.

\begin{figure}[ht]
\centering
\subfigure[]{\includegraphics[width=0.23\textwidth]{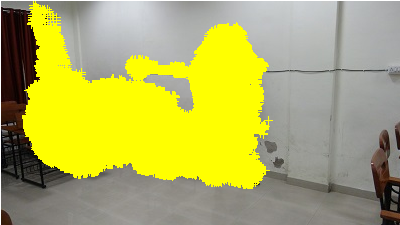}}
\subfigure[]{\includegraphics[width=0.23\textwidth]{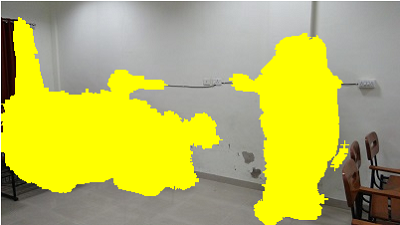}}
\caption{NRDC Correspondences for the salient regions of the images in Figure \ref{Fig:1}.} 
\label{Fig:3}
\end{figure}

We compute the dense correspondence using NRDC only on the salient regions of the two images \cite{hacohen2011non}. We use only those matched features which are matched with a confidence of more than $0.8$. These points are shown in Figure \ref{Fig:3}. This enables us to remove any false correspondences. We compute the difference between the matched points in both the images to obtain the motion vector between every matched pair of points. We know that the motion vector, thus estimated, comprises of both the camera motion and object motion (if any). We represent all these vectors as the rows of a matrix $W$. We compute the singular value decomposition of the matrix $W=UDV^{T}$. This decomposition will lead to two singular values as $W$ will be of size $N\times 2$, where $N$ is the number of matched points. The dominant motion in the matched points present in the static regions will be captured by one singular value. The motion in the matched points present in the dynamic regions will be captured by the other singular value.
\subsection{Identification of Static and Dynamic Features}
\begin{figure}[ht]
\centering
\subfigure[]{\includegraphics[width=0.23\textwidth]{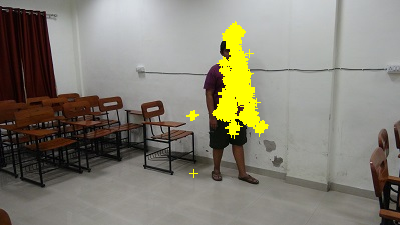}}
\subfigure[]{\includegraphics[width=0.23\textwidth]{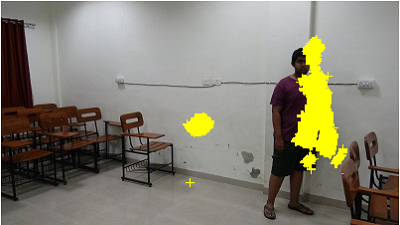}}
\caption{NRDC Correspondences for the interest points on the dynamic objects with outliers.} 
\label{Fig:4}
\end{figure}

We need to process this motion vector matrix $W$ generated using dense point correspondences in the salient regions of the scene in order to identity the static and dynamic regions. We shall now look at an optimal way to achieve this task. We can create a new matrix $\tilde{W}$ from $W$ by making the singular value corresponding to the static region zero. We use this approach proposed for object localization in \cite{Sai2015Dynamic}. This matrix will provide us coordinates of matched points only in the dynamic regions of the scene. The results are shown in Figure \ref{Fig:4}. One can observe some of the interest points not located on the dynamic object being detected. 

\begin{figure}[ht]
\centering
\subfigure[]{\includegraphics[width=0.23\textwidth]{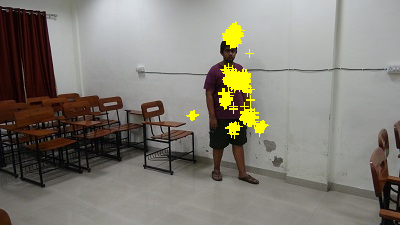}}
\subfigure[]{\includegraphics[width=0.23\textwidth]{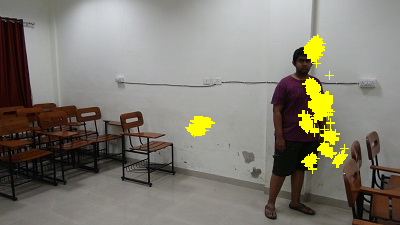}}
\caption{NRDC Correspondences corresponding to only the dynamic objects.} 
\label{Fig:5}
\end{figure}

Hence, we again decompose the matrix $\tilde{W}$ using SVD and retain only the motion vectors corresponding to the dominant singular value. This enables us to reduce some of the outliers (due to unwanted disturbances in the scene background) which are detected as belonging to the dynamic objects. Therefore, we have estimated the dense correspondence in the dynamic objects present in the scene as shown in Figure \ref{Fig:5}. We also notice that some of the interest points matched on the dynamic objects are also reduced. This will not affect us as we are concerned only about the envelope of the points rather than the points themselves. The next task is to segment the scene using these matched points present in the dynamic objects of the scene.

\subsection{Segmentation of Dynamic Objects}
\begin{figure}[ht]
\centering
\subfigure[]{\includegraphics[width=0.23\textwidth]{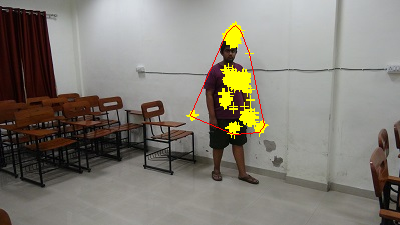}}
\subfigure[]{\includegraphics[width=0.23\textwidth]{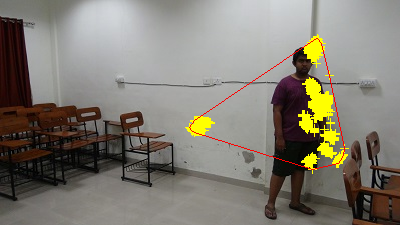}}
\caption{Convex hull fit over the dynamic interest points in Figure \ref{Fig:5}.} 
\label{Fig:6}
\end{figure}

We now have the interest points located on the dynamic objects present in the scene. Assuming there is only one dynamic object present in the scene, we fit a convex hull using all the dynamic interest point coordinates in both the images (\cite{graham1972efficient}, \cite{de2000computational}). The results are shown in Figure \ref{Fig:6}. This convex hull is then used to obtain minimum area bounding box using the algorithm proposed in \cite{freeman1975determining}. The bounding box may not be aligned with respect to the vertical and horizontal axes of the image. 

Let $(x_1,y_1),(x_2,y_2),(x_3,y_3),(x_4,y_4)$ be the four vertices of the rectangle obtained through the approach in \cite{freeman1975determining}. We would like to now compute the maximum and minimum values of the $x$ and $y$ coordinates of this rectangles as per the equations 1.
\begin{eqnarray}
x_{max}=\max\{x_1,x_2,x_3,x_4\}\\\nonumber
x_{min}=\min\{x_1,x_2,x_3,x_4\}\\\nonumber
y_{max}=\max\{y_1,y_2,y_3,y_4\}\\\nonumber
y_{min}=\min\{y_1,y_2,y_3,y_4\}
\label{Eq:1}
\end{eqnarray}

\begin{figure}[ht]
\centering
\subfigure[]{\includegraphics[width=0.23\textwidth]{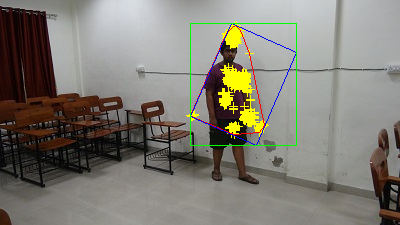}}
\subfigure[]{\includegraphics[width=0.23\textwidth]{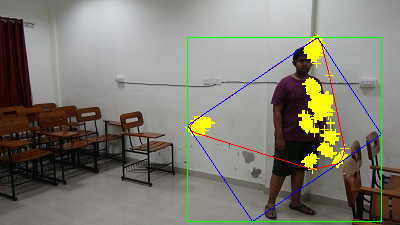}}
\caption{Rectangular minimal area bounding boxes fit using the convex hull in Figure \ref{Fig:6}.} 
\label{Fig:7}
\end{figure}

The new rectangle which is aligned with the vertical and horizontal axes of the given image is computed with the new coordinates estimated as $(x_{max}, y_{max}), (x_{max}, y_{min}), (x_{min}, y_{max}), (x_{min}, y_{min})$.  As most segmentation algorithms require the rectangular bounding box as an input, we perform this transformation of the vertices of the rectangle to obtain a new rectangle whose sides are aligned with the vertical and horizontal axes of the image. This process is illustrated in Figure \ref{Fig:7}. The blue bounding boxes which correspond to unaligned rectangles are aligned into green bounding boxes which are the aligned rectangles as evident in this figure.

\begin{figure}[ht]
\centering
\subfigure[]{\includegraphics[width=0.46\textwidth]{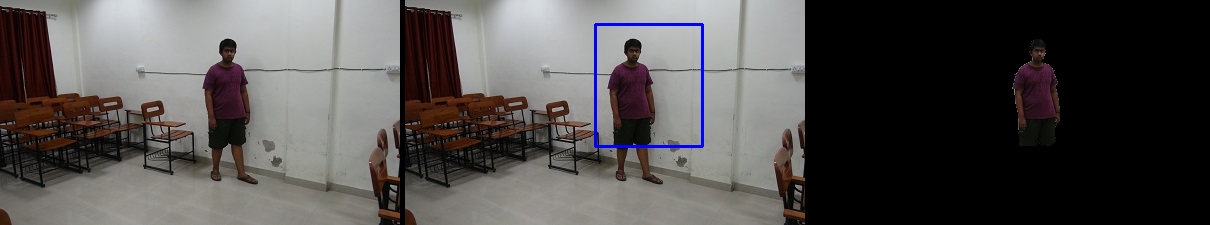}}
\subfigure[]{\includegraphics[width=0.46\textwidth]{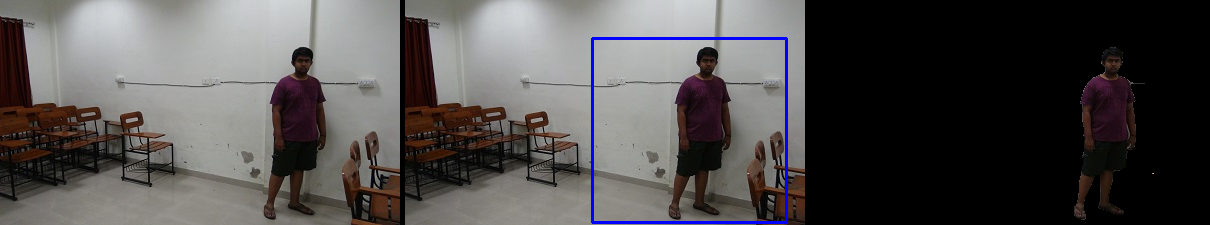}}
\caption{Images, estimated bounding boxes and the segmentation using GrabCut for the images in Figure \ref{Fig:1} (Dataset - 1). } 
\label{Fig:8}
\end{figure}

In case there are more number of dynamic objects present in the scene, we assume a prior knowledge of the number of such dynamic objects. We perform a K-means clustering in order to group the matched interest points corresponding to different dynamic objects. We fit a convex hull and estimate a bounding box using the matched interest points corresponding to each dynamic object separately. The final goal is to use the original images of the scene and the bounding boxes in order to segment out the dynamic objects. We achieve this task using the `GrabCut' algorithm which uses graph cut minimization in order to perform top-down segmentation \cite{rother2004grabcut}. The important observation is that the bounding box required for this algorithm has been obtained automatically without any human interaction. Therefore, the proposed approach is completely automatic and we are able to obtain the segmentation for all the dynamic objects present in the scene. The original images, estimated bounding boxes and the segmentation achieved are shown in Figure \ref{Fig:8}.

We evaluate the segmentation obtained using the proposed framework with the ground truth segmentation obtained manually. We would like to employ Jaccard similarity coefficient as the metric to analyse the efficacy of the automatic segmentation achieved using the proposed framework. Let $S_{e}$ be the segmentation mask estimated using the proposed framework and $S_{r}$ be the ground truth reference segmentation mask created manually. The Jaccard similarity coefficient between these two segmentation masks can be defined as shown in equation 2 \cite{tan2013introduction}. 
\begin{equation}
J(S_{e},S_{r})=\frac{|S_{e}\cap S_{r}|}{|S_{e}\cup S_{r}|}
\label{Eq:2}
\end{equation} 
 where $|.|$ denotes the cardinality of a given set.
\section{Results and Discussion}
In this section, let us look at the results of the proposed segmentation approach for different scenes captured using two images. We choose different scenes in the variety in the complexity with respect to the object and camera motions as well as the background clutter. We shall use the dataset created by us for this analysis as this problem is quite unique. We plan to release the dataset in public domain in future. In each case, we shall provide only the two images, estimated bounding box and the final segmentation result. The intermediate steps have already been discussed in detail in the previous section.

\begin{figure}[ht]
\centering
\subfigure[]{\includegraphics[width=0.46\textwidth]{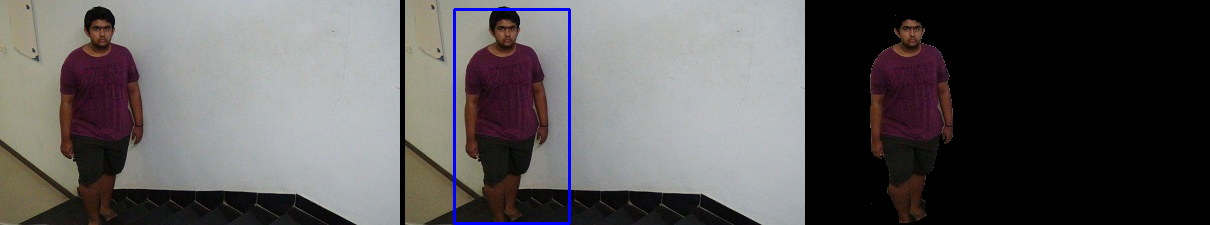}}
\subfigure[]{\includegraphics[width=0.46\textwidth]{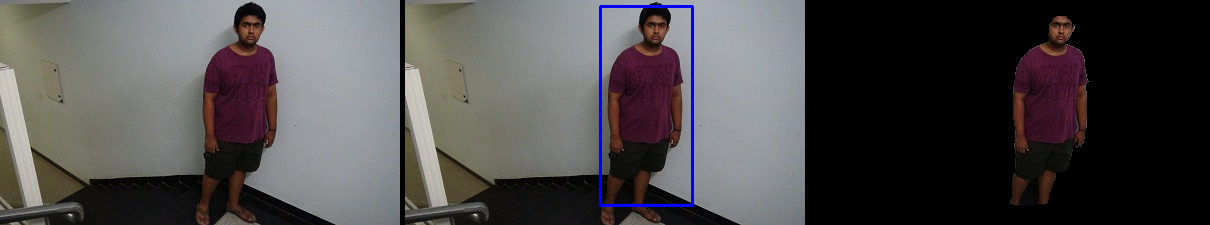}}
\caption{Results for Dataset - 2.} 
\label{Fig:9}
\end{figure}

Figure \ref{Fig:9} shows the images of a scene captured using a hand-held camera in which a person is in different positions. The background for this scene is simple and does not have much clutter as there are not much details. We can observe from the figure that we are able to estimate the bounding box around the person in both the images automatically. The GrabCut algorithm is able to segment the person under consideration quite effectively. 

\begin{figure}[ht]
\centering
\subfigure[]{\includegraphics[width=0.46\textwidth]{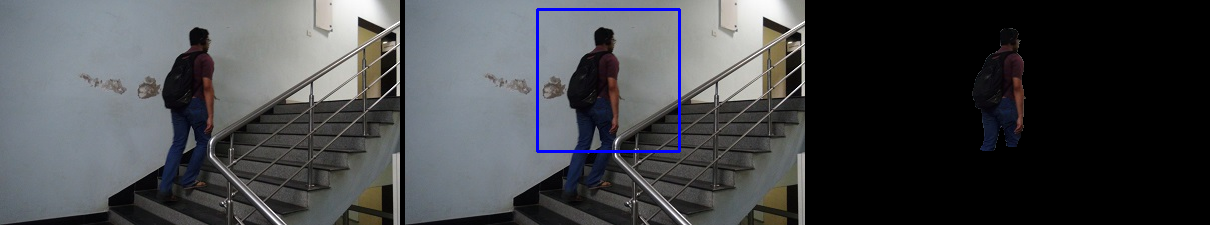}}
\subfigure[]{\includegraphics[width=0.46\textwidth]{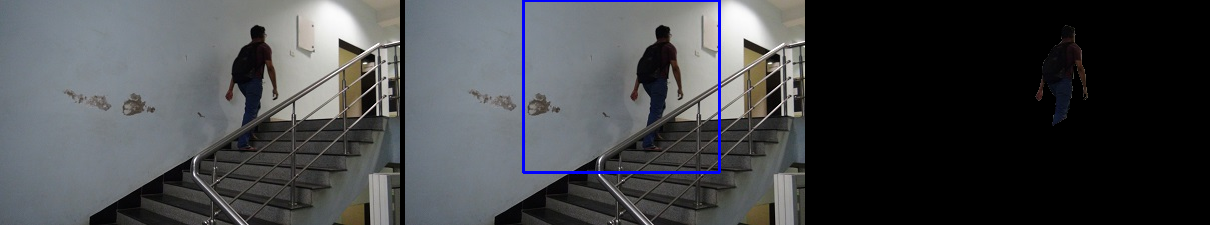}}
\caption{Results for Dataset - 3.} 
\label{Fig:10}
\end{figure}

Figure \ref{Fig:10} shows another dynamic scene captured using a hand-held camera. The person in the scene climbs up the stairs. The background clutter in the scene more complex compared to that of Figure \ref{Fig:9}. We can observe that the proposed approach leads to the estimation of bounding boxes which include significant amount of background along with the person who is changing the position. We can also observe that the segmentation of the person in this scene is also almost perfect using the GrabCut algorithm.

\begin{figure}[ht]
\centering
\subfigure[]{\includegraphics[width=0.46\textwidth]{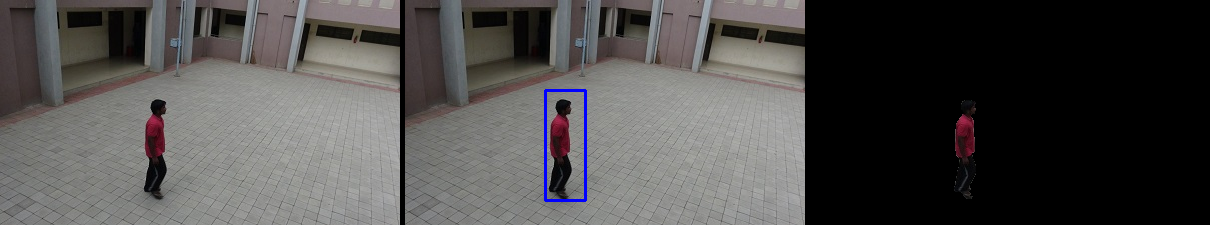}}
\subfigure[]{\includegraphics[width=0.46\textwidth]{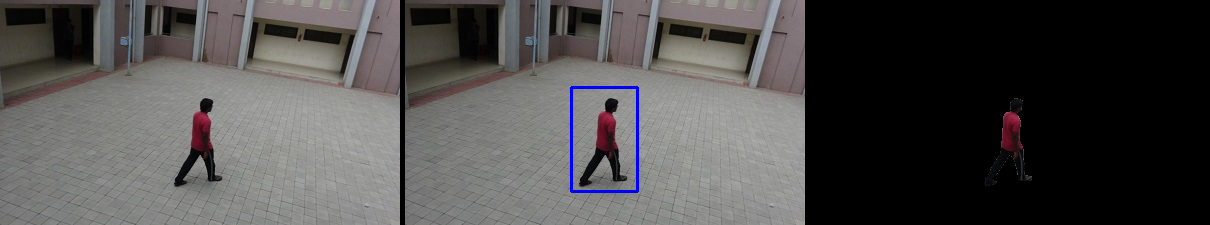}}
\subfigure[]{\includegraphics[width=0.46\textwidth]{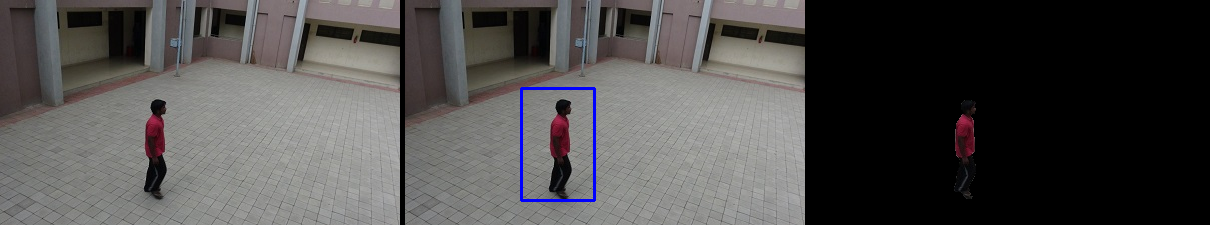}}
\subfigure[]{\includegraphics[width=0.46\textwidth]{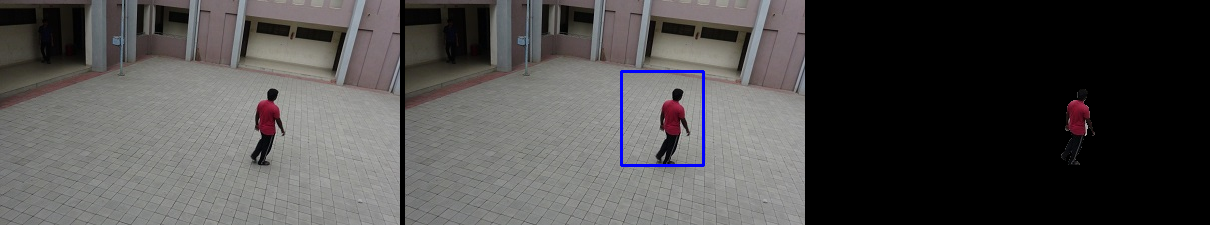}}
\caption{Results for Dataset - 4.} 
\label{Fig:11}
\end{figure}

Figure \ref{Fig:11} shows a set of three images of a dynamic scene captured using a hand-held camera. We take first image in Figure \ref{Fig:11} (a) as the reference image and try to segment the moving person from the other two images shown respectively in Figure \ref{Fig:11} (b) and (d). We can see from the first column that the first two images show a small amount of non-rigid motion of the person while the first and third images show that the person has undergone a non-rigid motion of large amount. From the second column, we could observe that we are able to estimate accurately the bounding box around the person for both the small and large displacement using the proposed approach. The last column shows that we are able to segment the person from all three images using GrabCut. This example shows that the proposed framework is able to segment both small as well as large displacements observed in a scene.

\begin{figure}[ht]
\centering
\subfigure[]{\includegraphics[width=0.46\textwidth]{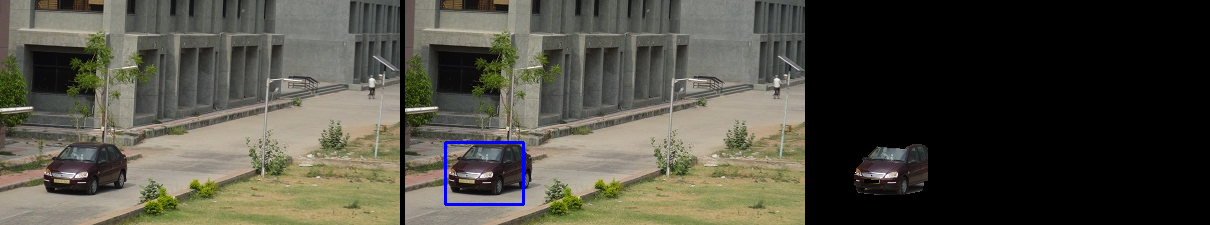}}
\subfigure[]{\includegraphics[width=0.46\textwidth]{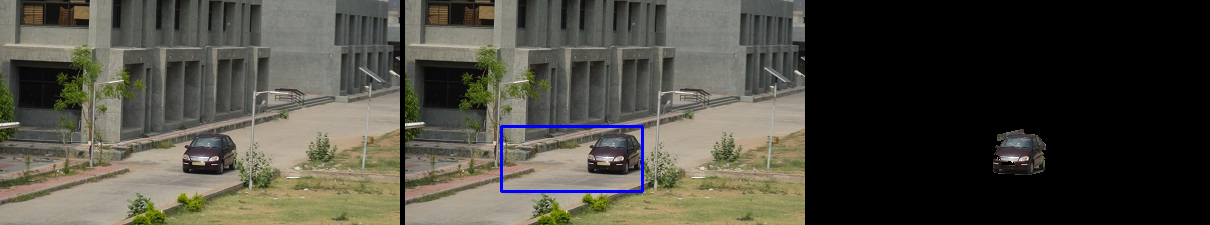}}
\subfigure[]{\includegraphics[width=0.46\textwidth]{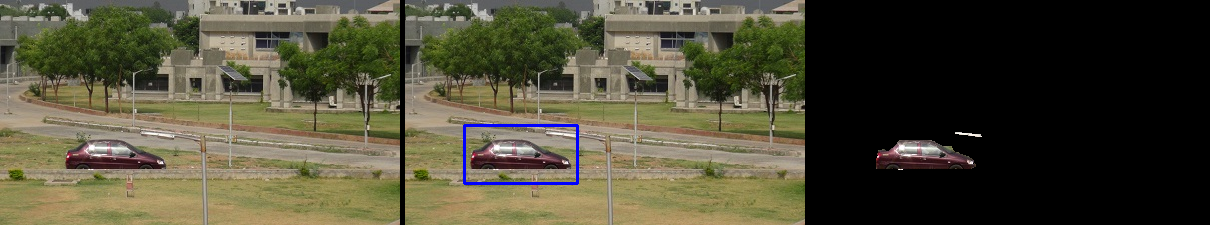}}
\subfigure[]{\includegraphics[width=0.46\textwidth]{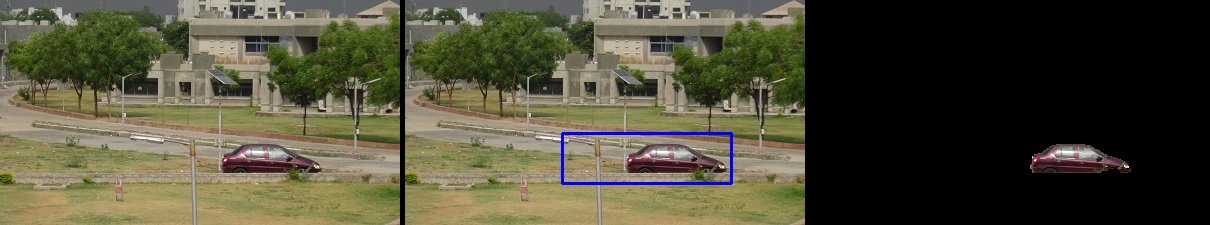}}
\caption{Results for Dataset - 5.} 
\label{Fig:12}
\end{figure}

We would now like to test the effectiveness of the algorithm on rigid motion on different planes with respect to the camera. Figure \ref{Fig:12}(a,b) show images of a car which is approaching towards the hand-held camera with a lot of background clutter. We are able to obtain the bounding boxes around the car in both the images as shown in the second column and we are also able to segment the car out. Figure \ref{Fig:12}(c,d) show images of a car which moves at a small angle with respect to the principal axis of the camera. We can observe from the second and third columns that we are able to correctly segment the cars in both the images using the proposed framework.

\begin{figure}[ht]
\centering
\subfigure[]{\includegraphics[width=0.46\textwidth]{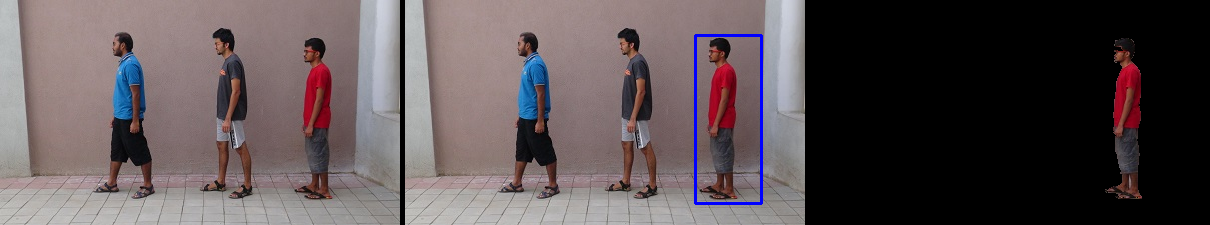}}
\subfigure[]{\includegraphics[width=0.46\textwidth]{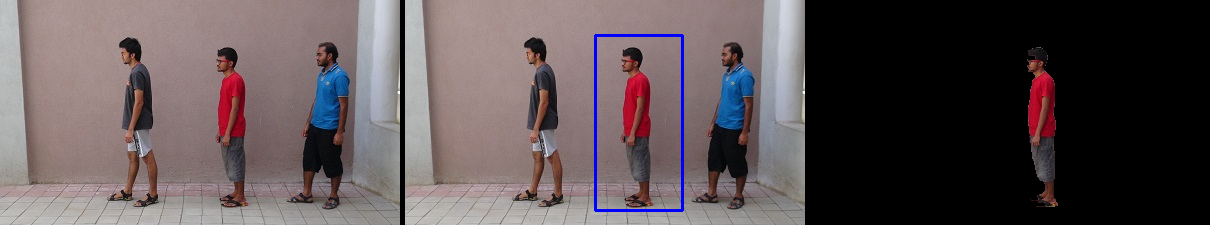}}
\subfigure[]{\includegraphics[width=0.46\textwidth]{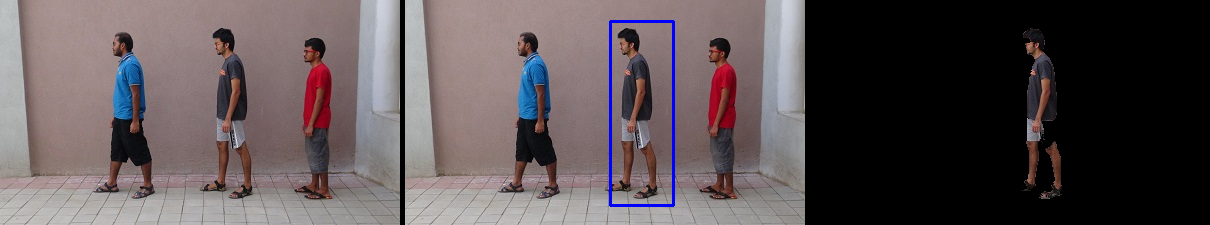}}
\subfigure[]{\includegraphics[width=0.46\textwidth]{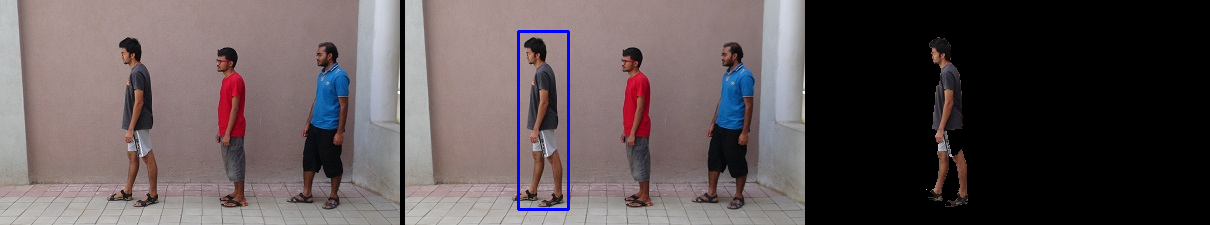}}
\subfigure[]{\includegraphics[width=0.46\textwidth]{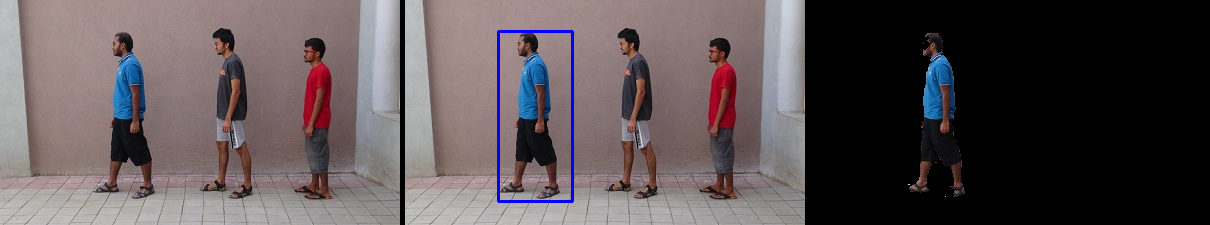}}
\subfigure[]{\includegraphics[width=0.46\textwidth]{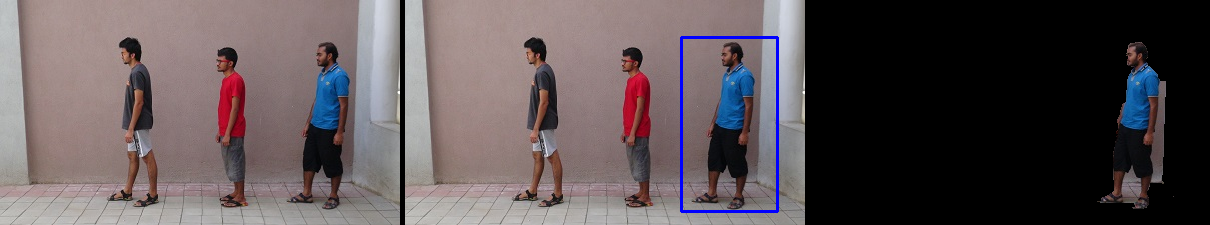}}

\caption{Results for Dataset - 6.} 
\label{Fig:13}
\end{figure}

The first column of Figure \ref{Fig:13} shows two images of a scene captured using a hand-held camera where multiple objects are in motion. We would like to segment out all the three persons who have changed their positions in the two images. One can observe from the second column that we are able to get the bounding box exactly for all the three persons using the proposed approach. The last column shows the results of the segmentation obtained for the three persons. This is an example to show that the proposed framework can be used to segment multiple objects effectively. The segmented results except the one in last row and last column are very good. The failure in segmenting the last image is due to the fact that we did not get a tight bounding box for the GrabCut to work on. To obtain these results we had assumed that the number of dynamic objects present in the scene is known in prior.

\begin{figure}[ht]
\centering
\subfigure[]{\includegraphics[width=0.46\textwidth]{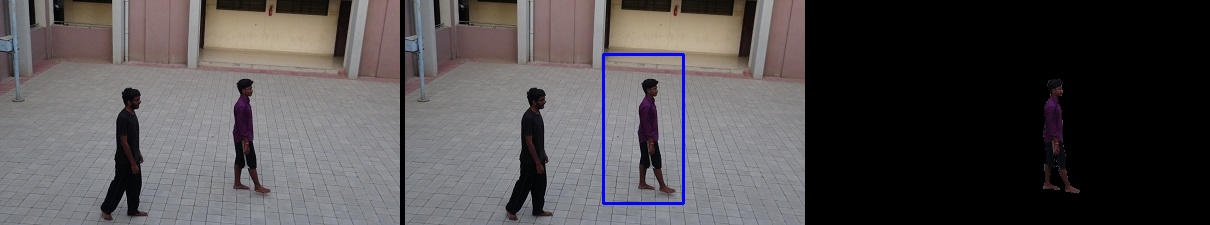}}
\subfigure[]{\includegraphics[width=0.46\textwidth]{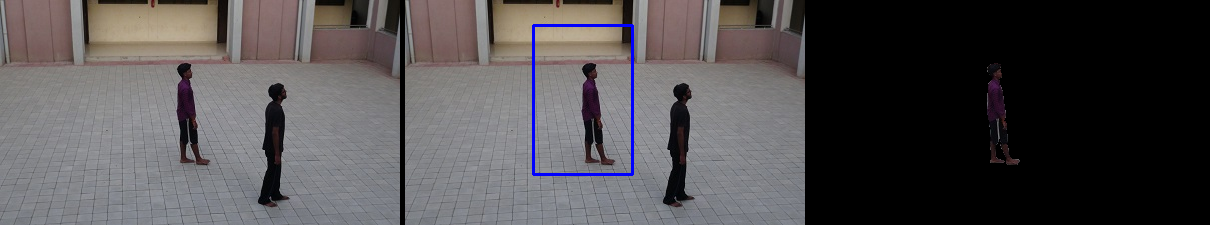}}
\subfigure[]{\includegraphics[width=0.46\textwidth]{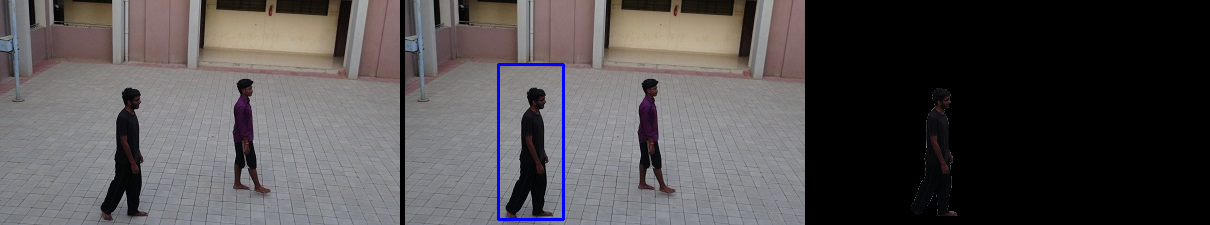}}
\subfigure[]{\includegraphics[width=0.46\textwidth]{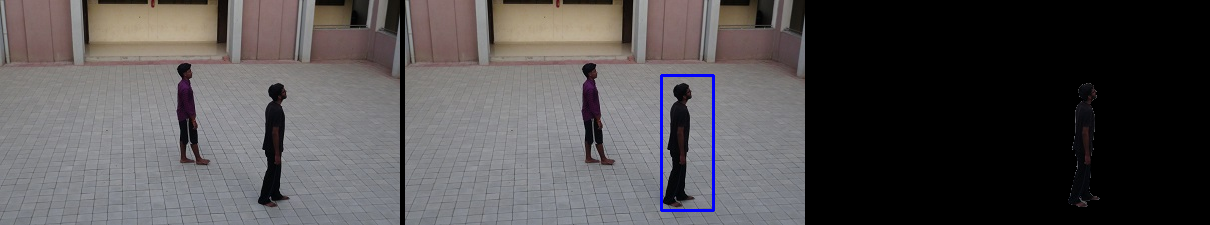}}
\caption{Results for Dataset - 7.} 
\label{Fig:14}
\end{figure}

Figure \ref{Fig:14} shows a scene in which two persons have changed their positions while capturing the two images. We have been able to detect the bounding box around the two persons in both the images and segment them out as observed in the second and third columns. One important observation is that the two persons do not occlude each other and hence we are able to achieve good segmentation for all the dynamic objects. Here we assumed that the number of objects present in the scene is known in prior.

The entire implementation of the proposed framework for segmenation took 40 seconds for typical images of size 225$\times$400 on a computer with Intel i5 core processor and 4GB RAM. We shall now look at the Jaccard similarity coefficient for all the results with respect to the manually marked segmentation masks. Table \ref{Table:1} shows the metric for various segmentation results obtained. We notice that we are able to achieve more than 0.75 for most of the segmentation masks using the proposed approach. In some cases we are able achieve even more than 0.9 which is more closer to the ground truth segmentation. In future, we would like to improve the method by addressing issues related to occlusion and the presence of fine structures on the object under consideration. The similarity index is very low for the dataset 8 as evident from the segmentation results as well.
\begin{table}
\begin{center}
\begin{tabular}{|c|c|c|c|}
\hline
 \textbf{Dataset \#} & \textbf{Figure \#} & \textbf{Image 1}& \textbf{Image 2}\\\hline
1 &Figure \ref{Fig:8}(a,b)& 0.98 & 0.84\\\hline
2 &Figure \ref{Fig:9}(a,b)& 0.92 & 0.86\\\hline
3 &Figure \ref{Fig:10}(a,b)& 0.74 & 0.90\\\hline
4 &Figure \ref{Fig:11}(a,b)& 0.90 & 0.88\\
  &Figure \ref{Fig:11}(c,d)& 0.91 & 0.91\\\hline
5 &Figure \ref{Fig:12}(a,b)& 0.74 & 0.81\\
  &Figure \ref{Fig:12}(c,d)& 0.89 & 0.92\\\hline
  &Figure \ref{Fig:13}(a,b)& 0.91 & 0.92\\
6 &Figure \ref{Fig:13}(c,d)& 0.75 & 0.85\\
  &Figure \ref{Fig:13}(e,f)& 0.90 & 0.73\\\hline
7 &Figure \ref{Fig:14}(a,b)& 0.92  & 0.92 \\
  &Figure \ref{Fig:14}(c,d)& 0.93  &  0.91 \\\hline

\end{tabular}
\end{center}
\caption{Jaccard similarity coefficient for the segmentation obtained using the proposed framework with respect to the ground truth segmentation.}
\label{Table:1}
\end{table}

\section{Conclusion}
We have proposed a novel framework for the segmentation of dynamic objects present in the scene from two images captured using a hand-held camera. The main novelty lies in the fact that we do not require any human interaction and are able to obtain the complete segmentation of the dynamic objects exhibiting even large motion. We have shown that the proposed approach is able to provide us very good segmentation of the object boundaries even for a large change in camera position while capturing the two images. We have also validated the segmentation obtained with the ground truth segmentation masks through Jaccard similarity coefficients for various results obtained and have discussed the effectiveness of the approach. The proposed approach can serve as a tool which provides useful cues to tasks such as object recognition and visual tracking which requires some initialization. 

We have shown that the proposed framework encounters problems in segmenting when multiple objects occlude each other and when there are fine structures present in an object. We also need to refine the bounding box estimation for these cases in order to obtain better segmentation. Segmenting multiple objects in a scene which are perspectively closer with respect to the camera is a challenge. We would like to improve the present framework in order to segment any number of moving object from two images captured using a hand-held camera. In future, we also want to relax the requirement of the prior knowledge of the number of dynamic objects present in the scene. Another interesting future idea could be to find out the smallest and largest object motion which the proposed approach could detect and segment. We would like to explore more on these possibilities of research and make this novel approach for segmentation more robust in future.


%

%
%
%




\end{document}